%% file: emnlp2021.tex
\pdfoutput=1
\documentclass[11pt]{article}

\usepackage[]{acl}

\usepackage{times}
\usepackage{latexsym}

\usepackage[T1]{fontenc}
\usepackage[utf8]{inputenc}
\usepackage{microtype}
\usepackage{booktabs}
\usepackage{graphicx}
\usepackage{multirow}
\usepackage{footmisc}
\usepackage{amsfonts}
\usepackage{tabularx}
\usepackage{verbatim}
\usepackage{subfigure}
\usepackage{enumitem}

\usepackage{amsmath}
\DeclareMathOperator*{\argmax}{arg\,max}
\definecolor{mypink}{RGB}{219, 48, 122}

\title{Lexical Knowledge Internalization for Neural Dialog Generation}

\author{
 Zhiyong Wu$^{1,2}$~\thanks{\, The majority of this work was done while the first author was interning at Tencent AI Lab.}\,, Wei Bi$^{3}$~\thanks{\, Corresponding author}, Xiang Li$^{4}$, Lingpeng Kong$^{1,2}$, Ben Kao$^{1}$\\
 $^{1}$The University of Hong Kong, $^{3}$Tencent AI Lab, \\
 $^2$Shanghai AI Lab, $^{4}$East China Normal University \\
 $^{1}$\{zywu,lpk,kao\}@cs.hku.hk, $^{3}$victoriabi@tencent.com, $^{4}$xiangli@dase.ecnu.edu.cn\\
}

\begin{document}
\maketitle
\begin{abstract}
We propose knowledge internalization (KI), which aims to complement the lexical knowledge into neural dialog models. Instead of further conditioning the knowledge-grounded dialog (KGD) models on externally retrieved knowledge, we seek to integrate knowledge about each input token internally into the model's parameters. To tackle the challenge due to the large scale of lexical knowledge, we adopt the contrastive learning approach and create an effective token-level lexical knowledge retriever that requires only weak supervision mined from Wikipedia. We demonstrate the effectiveness and general applicability of our approach on various datasets and diversified model structures.

\end{abstract}

\input{1-introduction}

\input{2-related}
\input{3-method}
\input{4-retrieval}
\input{5-exp}

\input{6-analysis}

\section{Conclusion}
We propose knowledge internalization (KI), which aims to incorporate the lexical knowledge into neural dialog models. Models with KI can generate informative and diverse responses without explicitly conditioning on external knowledge.
To provide the fine-grained knowledge needed in KI, we also build an effective token-level lexical knowledge retriever that contextually align tokens in a sentence to their related knowledge.
We show the effectiveness and general applicability of KI by evaluating KI on various datasets and diversified model structures.

\section{Acknowledgement}
This project is supported by the Tencent AI Lab Rhino-Bird Focused Research Program, the Shanghai Committee of Science and Technology, China (Grant No. 21DZ1100100). This research is also partly supported by the HKU-TCL Joint Research Center for Artificial Intelligence fund (Project 200009430).

\bibliographystyle{acl_natbib}
\bibliography{anthology,custom}
\input{9-appendix}
\end{document}

%% file: 1-introduction.tex
\section{Introduction}

\begin{figure}[ht]
    \centering
    % \resizebox{\linewidth}{!}{
    \subfigure[An utterance-response-knowledge triple. \label{fig:intro_a}]{
    \includegraphics[width=\linewidth]{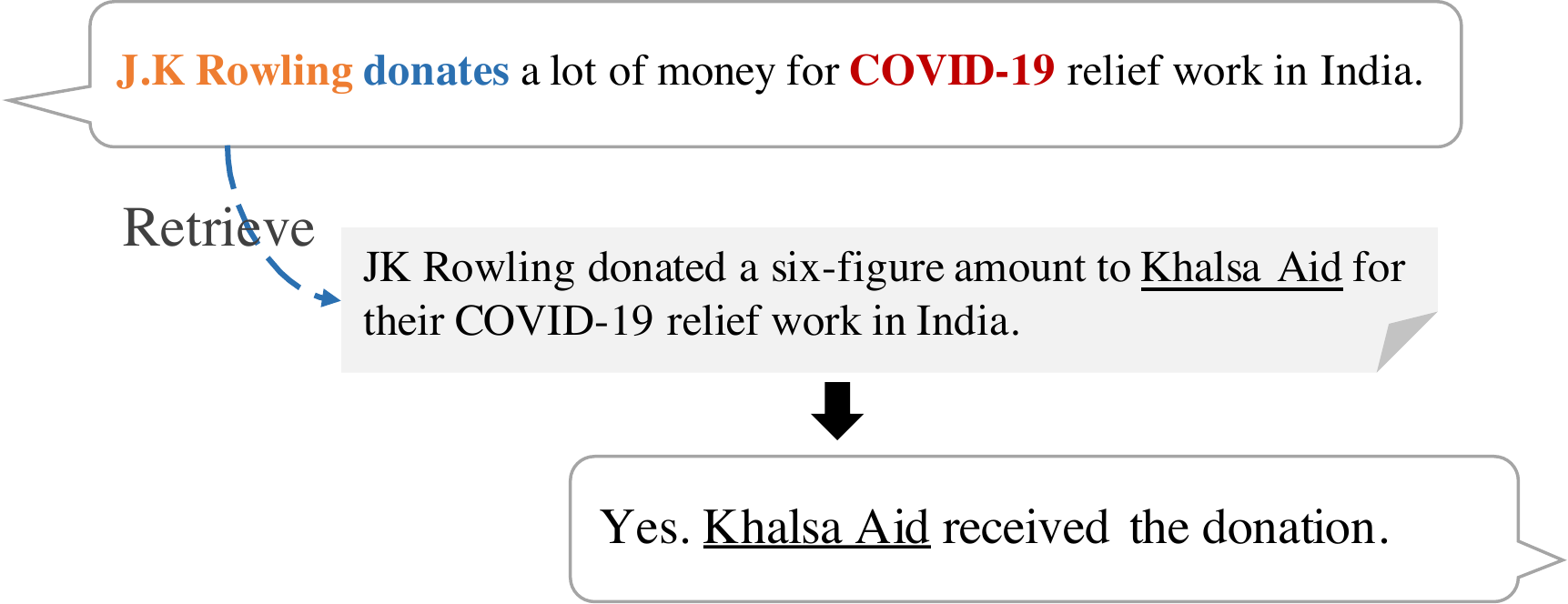}
    }
    \subfigure[Responses based on different lexical knowledge. \label{fig:intro_b}]{
    \includegraphics[width=\linewidth]{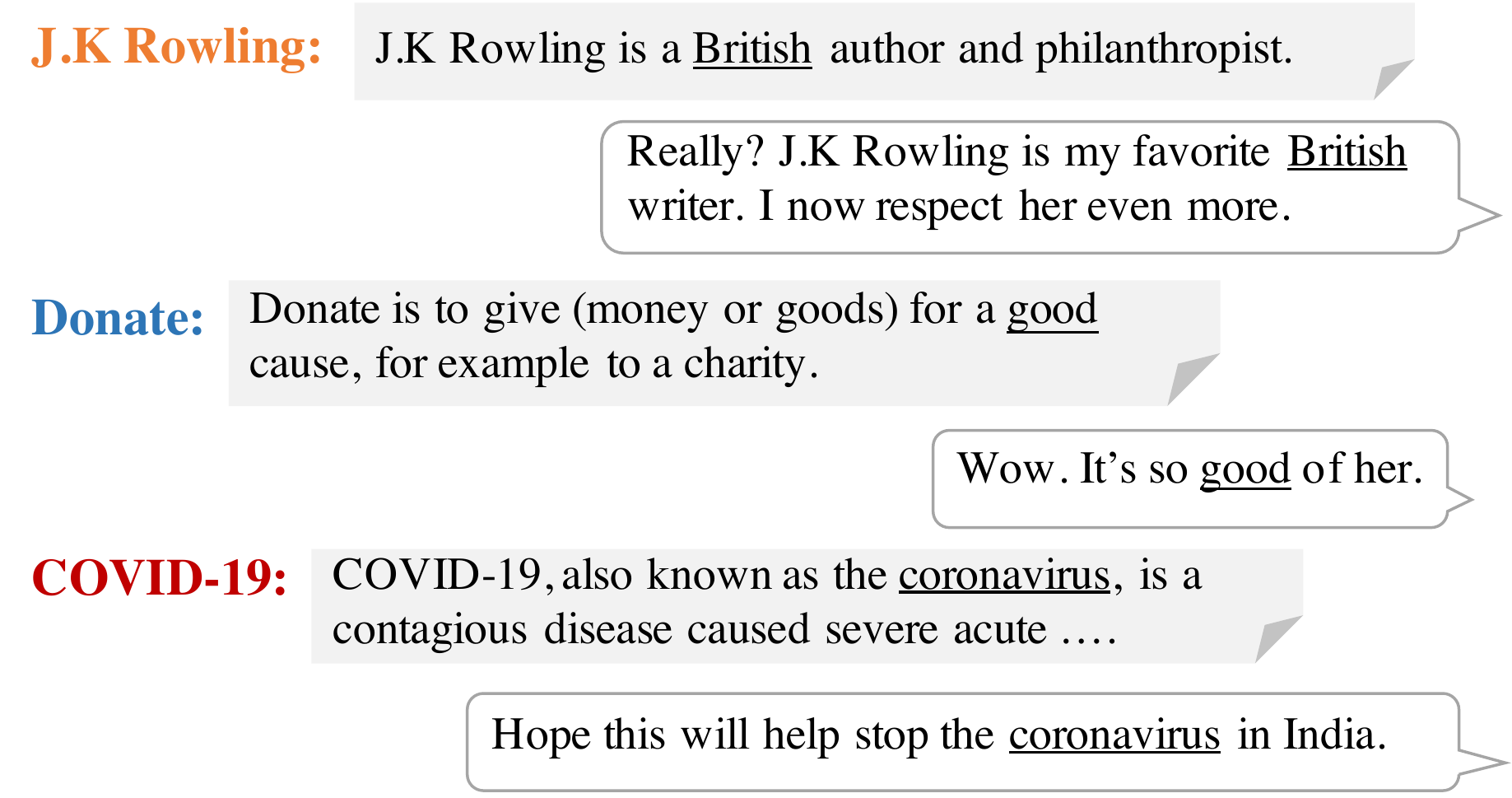}
    }
    \caption{(a) An exemplary KGD data sample with an utterance (top), a response (bottom), and a sentence-level knowledge (middle). (b) A list of lexical knowledge (in grey rectangle) related to words from the utterance in (a), and the potential responses (in white speech balloon) people would make given that knowledge.}
    \label{fig:example}
\end{figure}

Vacuous responses~\cite{li-etal-2016-diversity,ghazvininejad2018knowledge}, such as, \textit{I don't know}, are commonly observed in end-to-end neural dialog models~\cite{shang-etal-2015-neural,sordoni-etal-2015-neural}. This is mostly because these models ignore the knowledge that resides in people's minds during a conversation.  %due the ignorance of human knowledge in conversations.
To bridge the gap, many existing works~\cite{moghe-etal-2018-towards,dinan2018wizard} have attempted to condition the dialog model on external knowledge, either a sentence or a paragraph, retrieved based on the utterance and/or previous context.
This curates datasets with utterance-response-knowledge triples (see Fig~\ref{fig:intro_a}). 
These knowledge-grounded dialog (KGD) models, despite demonstrated effectiveness, suffer from two major problems. 

First, equipping models with sentence-level knowledge alone will limit responses' informativeness and diversity. As shown in Fig~\ref{fig:intro_a}, with the knowledge retrieved giving the utterance, a KGD model can relate \textit{J.K Rowling} to \textit{Khalsa Aid}. 
However, retrieval based solely on sentence embeddings will result in ignorance of lexical knowledge associated with individual tokens. % 
%\bi{the above sentence is also a bit strange.}
In this example, the knowledge about \textit{J.K Rowling}, \textit{COVID-19}, \textit{donates}, and \textit{India}, is ignored during the retrieval, due to the semantic gaps between those lexical knowledge sentences (see Fig~\ref{fig:intro_b}) and the utterance. This makes it rather difficult (if not impossible) for the model to generate responses carrying relevant information as shown in Fig~\ref{fig:intro_b}. 

Second, retrieving knowledge for open-domain dialogs during inference incurs heavier computation, often involving similarity search over tens of millions of passages~\cite{petroni2021kilt}. 

Existing systems~\cite{zhao-etal-2020-knowledge-grounded,zheng-etal-2020-difference} alleviate this problem relying on pre-selecting a small candidate set based on TF-IDF~\cite{schutze2008introduction}, in sacrifice of the diversity and the accuracy of the retriever. 
Directly conditioning the dialog model on the retrieved text, these models 
% are prone to errors as the generation heavily depends on 
are easily effected by the quality of the constructed candidate set and are thus prone to errors~\cite{dinan2018wizard,Kim:2020:Sequential,zhao-etal-2020-knowledge-grounded}.
 
In this work, we propose to complement the lexical knowledge into neural dialog models by \textbf{K}nowledge \textbf{I}nternalization (\textbf{KI}), a training approach based on contrastive learning~\cite{hadsell2006dimensionality}.
The central idea of KI is to integrate more fine-grained lexical knowledge about each input token internally into model parameters (e.g., word embeddings), rather than further conditioning the model on externally retrieved knowledge (e.g., directly copy and/or modify tokens from external knowledge when decoding).
Our research contributions include:
\begin{itemize}[wide=0.5\parindent,noitemsep, topsep=0.5pt]
\item a novel training objective (KI; \S\ref{sec:ki-loss}) that infuses lexical semantics into word representations. With the knowledge internalized into the contextualized representation of every token, a dialog model can generate informative and diverse responses \emph{without} engaging an external knowledge retrieval module during inference time, thus making the inference more efficient (\S\ref{subsec:base_experiment});

\item  %a contrastive learning approach and 
an effective token-level lexical knowledge retriever (\S\ref{sec:retrieval}) trained with weak supervision to contextually align tokens in dialog corpora to their related and possibly different knowledge (Appendix~\ref{app:retrieval}).
%mined from Wikipedia. 
%During training of a dialog model, the retriever assigns each utterance token to one item in the curated lexical database that consists of 6.4 million sentences. This provides fine-grained knowledge to boost the performance of the dialog model;
\item  a demonstration of the effectiveness and general applicability of KI with extensive experiments on diversified dialog models and on three benchmark datasets:
DailyDialog~\cite{li-etal-2017-dailydialog}, Wizard of Wikipedia~\cite{dinan2018wizard}, and Commonsense Reddit Dataset~\cite{zhou2018commonsense}. The implementation of our model can be found at \url{https://github.com/LividWo/KI}.
% with or without external knowledge retrieval module. 
\end{itemize}

%% file: 2-related.tex
\section{Related Work}
\label{sec:related}

To address the vacuous responses problem in neural dialog models, researchers propose to ground dialogs on real world knowledge and construct new corpora that contain utterance-response-knowledge triples. 
Specifically, responses are grounded to external knowledge derived from different knowledge sources~\cite{zhou2018commonsense, liu-etal-2018-knowledge, wu-etal-2019-proactive, dinan2018wizard, moghe-etal-2018-towards, ghazvininejad2018knowledge, mostafazadeh-etal-2017-image, meng2020openvidial, zhang-etal-2020-grounded}. Among different sources, textual knowledge~\cite{dinan2018wizard, parthasarathi-pineau-2018-extending, qin-etal-2019-conversing} receives the most attention as it is easy to obtain and scale.
%knowledge graphs~\cite{zhou2018commonsense,liu-etal-2018-knowledge,wu-etal-2019-proactive},  encyclopedia~\cite{dinan2018wizard,parthasarathi-pineau-2018-extending,qin-etal-2019-conversing}, movie scripts~\cite{moghe-etal-2018-towards}, review posts~\cite{ghazvininejad2018knowledge}, and visual environment~\cite{mostafazadeh-etal-2017-image, meng2020openvidial}. %These models are trained on utterance-response-knowledge triples, in which there is a strong connection between the response and associated knowledge. Such that by optimizing the cross entropy over the response, the model can achieve knowledge grounding.
However, the construction of knowledge-grounded datasets is costly and time-consuming. %Recent works thus propose to build more practical systems (without assuming a given knowledge) by enhancing KGD models with an extra knowledge selection component
To build a more practical system without assuming a given knowledge, recent studies enhance KGD models with an extra knowledge selection component~\cite{dinan2018wizard,Kim:2020:Sequential,zheng-etal-2020-difference,zhao-etal-2020-knowledge-grounded}. 

Most existing KGD models can be viewed as \textit{models with externalized knowledge}, where knowledge is explicitly used as part of the model input. The principle behind these models is to copy words and/or modify sentences from external knowledge when generating responses~\cite{wu-etal-2020-diverse,zhu2017flexible,zhao2019low}. 
Our KI, on the other hand, does not explicitly present knowledge to dialog models for reading and/or copying. Instead, we inject and store external knowledge into models' parameters and encourage models to elicit the encoded knowledge during generation. 

The idea of knowledge internalization has also been explored in language modeling. Factual knowledge~\cite{zhang-etal-2019-ernie,sun2020ernie,liu2020k}, visual knowledge~\cite{tan-bansal-2020-vokenization} and syntactic knowledge~\cite{kuncoro-etal-2020-syntactic} have been injected into language models (LMs) and shown great promise in improving the performance of downstream tasks. KI differs from those knowledge-enhanced LMs in two aspects: (i) KI can be trained end-to-end with dialog models, while applying LMs on dialog generation often requires multiple rounds of pre-train and fine-tune. (ii) KI is lightweight that barely introduces extra parameters to the dialog model while applying LMs usually introduces hundreds of millions of extra parameters.

%% file: 3-method.tex
\section{Knowledge Internalization for Neural Dialog Models}
\label{sec:method}
In this section, we illustrate how to train a dialog model with knowledge internalization. To infuse more fine-grained lexical knowledge to a neural dialog model, we assume a dialog corpus where each token is aligned with relevant knowledge (we will discuss the construction of such a corpus in \S\ref{sec:retrieval}). In particular, for an input sentence $X$ in the corpus, we assume each token $x_i \in X$ is associated with a corresponding descriptive sentence $K_i$.

\subsection{Preliminary}

Given an utterance-response pair $(X,Y)$, where $X=\{x_1, x_2, \ldots, x_n\}$ and $Y=\{y_1, y_2, \ldots, y_m\}$, neural dialog models generally minimize the negative log-likelihood loss:
\begin{equation} \label{eq:nll}
\mathcal{L}_{\mathrm{NLL}}(X,Y) = - \sum_{i=1}^m log \mathcal{P}(y_i), 
\end{equation}
where $\mathcal{P}(y_i)= \mathcal{P}(y_i| y_{\textless i}, X)$ is the probability of generating the $i$-th response token $y_i$ given the utterance $X$ and other tokens generated in previous steps $y_{\textless i}=\{y_1, y_2, \ldots, y_{i-1}\}$. $\mathcal{P}(y_i)$ is generally modeled by a sequence-to-sequence model~\cite{sutskever2014sequence}, which consists of an encoder and a decoder. 
The encoder represents $X$ as a sequence of hidden vectors $H(X)={h_1, h_2, ..., h_n}$, where each $h_i$ is a low-dimensional representation of the token $x_i$. The decoder generates $y_i$ based on $H(X)$ and $y_{\textless i}$, 
often with the attention mechanism~\cite{bahdanau2014attention}.

\subsection{Knowledge Internalization Loss}
\label{sec:ki-loss}

Given a dialog corpus with token-level knowledge as discussed above, we now introduce a new training task: knowledge internalization (\textbf{KI}).
In KI, we seek to boost dialog models by internalizing lexical knowledge into each token's representation.
In particular, each token $x_i$ and its associated knowledge $K_i$ are first mapped into a low-dimensional space.
We then adopt contrastive learning to shorten the distance between $x_i$ and $K_i$ in the space while enlarging that between $x_i$ and other irrelevant knowledge.

Note that for each $x_i \in X$, dialog models' encoder can embed it into a contextualized representation $h_i$.
Therefore,
we only need an extra knowledge encoder to represent $K_i$ as $g(K_i)$ (details will be given in \S~\ref{sec:retrieval-encoder}).
After $h_i$ and $g(K_i)$ are computed,
we calculate the similarity between $x_i$ and $K_i$ by the inner product:
\begin{equation}\label{eq:relevance}
   s(x_i, K_i)=f_1(h_i)^{\top}f_2(g(K_i)), 
\end{equation}
where $f_1$ and $f_2$ are the functions that map the $h_i$ and $g(K_i)$ into the same vector space and normalize them.
% where 
% \begin{equation} \label{eq:mlp_norm}
%     f_1(h_i)=\frac{m l p_1(h_i)}{\|m l p_1(h_i)\|},
% \end{equation}
% \begin{equation} \label{eq:mlp_norm}
%     f_2(g(K_i))=\frac{m l p_2(g(K_i))}{\|m l p_2(g(K_i))\|}.
% \end{equation}
%To construct the a negative sample pair $(x_i, K^{-}_i)$,
%we perform an in-batch randomly sample of $K^{-}_i$, where $K^{-}_i \ne K_i$.

For each $(x_i, K_i)$ pair, we randomly sample an in-batch unrelated knowledge $K^{-}_i$ associated with other input sentences, where $K^{-}_i \ne K_i$, to construct a negative sample pair $(x_i, K^{-}_i)$ in contrastive learning.
%\bi{the in-batch unrelated knowledge here is not clear. u mean knowledge from other X or from other X union $x$ from current X.}
Finally, the objective function of KI is defined by the contrastive loss between positive and negative sample pairs:
\begin{equation}\label{eq:ki}
    \mathcal{L}_{KI}(X) \!\!=\!\!\!\sum_{i=1}^{n} \max\!{\left\{ 0, m\!-\!s(x_i, K_i)\!+\!s(x_i, K^{-}_i) \right\}},
\end{equation}
where $m$ denotes the margin.

\subsection{Knowledge-internalized Neural Dialog Model}
We now illustrate how to deploy KI on a neural dialog model.
We use a sequence-to-sequence dialog model based on Transformer~\cite{vaswani2017attention} as an example.
The original model is trained to minimize the negative log-likelihood loss of response tokens, i.e.,  $\mathcal{L}_{\mathrm{NLL}}(X,Y)$ (see Eq.~\ref{eq:nll}). %, as illustrated in Fig.~\ref{fig:overview} (right side). 
We can conveniently integrate KI into the model by reusing the contextualized representations generated by the model's encoder. 
%In particular, we aim to learn a hidden embedding space (gray rectangle with dotted line) where the word ``J.K Rowling'' is close to knowledge ``J.K Rowling is a British author and philanthropist.'', while being far away from ``A bichon is a distinct type of toy dog''. 
The training objective of a knowledge-internalized dialog model can then be formulated as:
\begin{equation}
    \mathcal{L} = \mathcal{L}_{\mathrm{NLL}}(X,Y) + \lambda \mathcal{L}_{KI}(X)
\end{equation}
where $\lambda$ is a hyperparameter.
%to control the interpolation. 
%
Note that the token-level knowledge is only required during training to compute $\mathcal{L}_{KI}(X)$. At the inference time, those relevant knowledge is no longer required as they have been \emph{internalized} into model by KI, making inference more efficient. 

%% file: 4-retrieval.tex
\section{Retrieval of Token-level Lexical Knowledge}
\label{sec:retrieval}
%In the previous section, we have presented how to use token-level knowledge to compute the KI loss in dialog models. our solution to construct training corpora required in \S~\ref{sec:method} -
In this section, we present how to train an effective retriever to mine knowledge for each token in the dialog corpora. Given a dialog utterance $X=\{x_1, x_2, \ldots, x_n\}$, the trained retriever will retrieve a relevant knowledge $K_i$ for each token $x_i$ in $X$. The constructed token-knowledge alignments can then be used to train a knowledge-internalized neural dialog model as in \S~\ref{sec:method}. 
% However, 
% there is no existing token-knowledge annotations that suits our need. In this section, we develop a model that retrieves such fine-grained knowledge from Wikipedia.
\subsection{Training Data Collection}
To train such a retriever, we need a corpus with token-level knowledge annotated. However, to our best knowledge, no human annotated data exist and it is prohibitively expensive to build one. We therefore seek to train the retriever with distant supervision.
A straight-forward solution is to align the noun words in an utterance to certain knowledge graph triples using entity linking tools~\cite{shen2014entity}. The problem of that is it can only cover about 15\% words in human conversations~\cite{biber2000longman}. 

%To resolve this challenge, we propose to mine token-knowledge alignments from Wikipedia. We observe that the first sentence of each Wikipedia article is always a declarative sentence that gives a high-level definition about the concept discussed in the article. Thus it can serve as the source of lexical knowledge. We then create sentence-knowledge alignments by linking all sentences in a Wikipedia article to its first sentence. Inspired by \citet{tan-bansal-2020-vokenization}, we then 
%break a sentence-knowledge mapping to a set of token-knowledge mappings by pairing all words in the sentence to the same knowledge.
%Given such token-knowledge mappings mined from Wikipedia, we are now being able to train a retriever that assigns relevant knowledge for each token in dialog corpus.
To address this issue, we propose to mine token-knowledge distant annotations from Wikipedia. %use the alignment between a token and a (representative sentence in) Wikipedia article as the lexical knowledge.
In each Wiki article, the first sentence $S=\{s_1, s_2, ..., s_n\}$ is mostly declarative that gives a high-level summary on the topic of the article. Thus this sentence can used as a lexical knowledge item, denoted as $K$ (note that $K$ and $S$ refer to the same sentence here).  Inspired by \citet{tan-bansal-2020-vokenization}, we then further associate every token in the sentence with this knowledge item. These constructed alignments (e.g., $(s_i,K)$) can then be used to train a token-level knowledge retriever.
%We then create a knowledge base with 6.4 million knowledge items by taking the first sentence from each Wiki article.

\subsection{Training of Retriever}
\label{sec:retrieval-encoder}
% a token-knowledge supervision $(s_i,K)$ and the context $S$, 
The core of the retriever's training is to learn a scoring function $r(s_i|S, K)$ to measure the relevance between a token $s_i$ and a knowledge item $K$, giving $s_i$'context $S$. Similar as Eq.~\ref{eq:relevance}, we implement the scoring function $r(s_i|S, K)$ as the inner product between $s_i$'contextualized token representation $f(h_i)$ and the knowledge representation $f(g(K))$. 
Here, we use a pre-trained $\text{BERT}$~\cite{devlin2019bert} model to obtain $h_i$; we apply another pre-trained BERT model to encode knowledge $K$ and further generate $g(K)$ with an average-pooling operator. Two BERT models will be fine-tuned with the retriever. 

Our training objective is to maximize the relevance score of aligned token-knowledge pairs while minimizing that of unaligned pairs. We also adopt the hinge loss similar as in Eq~\ref{eq:ki} by replacing $x_i$ in the dialog corpus to $s_i$ in the constructed token-knowledge pairs.

\subsection{Mining Token-level Lexical Knowledge}
\label{sec:biases}
Once the retriever is trained, we can use it to mine token-level lexical knowledge required in KI.
We first construct a candidate knowledge base $\mathcal{K}$ that consists of 6.4 million knowledge items (first sentence) extracted from Wikipedia articles.
Given a dialog utterance $X=\{x_1, x_2, \ldots, x_n\}$, we retrieve a lexical knowledge $K_i$ for each token $x_i$ by searching for the knowledge item that has the largest relevance score with $x_i$. 
\begin{equation}
K_i = \argmax_{K \in \mathcal{K}} r(x_i|X, K)
\end{equation}
To improve the retrieval results, we further employ two useful strategies: (i) \textit{Stopword Masking}, where we discard knowledge associated with stopwords; (ii) \textit{Exact Matching}, where if an utterance token exactly matches the title of a Wikipedia article, 
we will directly return the first sentence of this article as the retrieval result.

The retrieval process has two properties that can significantly improve dialog corpora's knowledge coverage. First, the retrieval is \textit{contextualized} such that a token can be aligned to different knowledge items when it occurs in different contexts. Second, the retrieval is at token-level that enables us to associate each dialog sentence with multiple knowledge items (See Appendix~\ref{app:retrieval}). 

%% file: 5-exp.tex
\section{Experimental Setups}
In this section, we present the datasets and metrics used for evaluation. 

\paragraph{Datasets}
We use three datasets from various domains (statistics in Appendix~\ref{app:dataset}).
The first one is \textit{DailyDialog}~\cite{li-etal-2017-dailydialog}, a multi-turn dialog benchmark that contains daily dialogs recorded as utterance-response pairs. 
However, there is no knowledge associated with the dialogs in DailyDialog, making it difficult to evaluate the informativeness of generated responses.
%Despite the depth of these daily dialogs is limited, we are interested in the informativeness of generated responses when equipping the model with token-level knowledge.
% the dataset lacks knowledge associated with conversations,
% which increases the difficulty in evaluating the informativeness of generated responses.
To fully illustrate the strength of KI, we further consider two knowledge-grounded datasets:
(i) \textit{Wizard of Wikipedia (WoW)}~\cite{dinan2018wizard}, a multi-turn dataset that contains utterance-response-knowledge triples. For each dialog, a sentence retrieved from Wikipedia is selected to guide response generation. 
WoW contains two test sets: Test Seen/Unseen, where the latter includes topics that never appear in Train and Valid set.
(ii) \textit{Commonsense Reddit Dataset (CRD)}~\cite{zhou2018commonsense}: a weakly knowledge-grounded single-turn dataset. 
Each dialog in the dataset is paired with at least one triple automatically extracted from ConceptNet~\cite{speer2017conceptnet}.
%This ensures that the subject/object of the triple appears in the conversation. 

\paragraph{Metrics}
We conduct both automatic evaluation and human annotations.
For automatic evaluation, we evaluate the generated responses from three perspectives~\footnote{For PPL and \%safe, smaller is better, while for all other metrics, larger is better.}:

\noindent$\bullet$ \textit{Appropriateness}: we employ \emph{Perplexity} (PPL), corpus-level BLEU-4~\cite{papineni-etal-2002-bleu}
%}~\footnote{https://github.com/moses$\text{-}$smt/mosesdecoder/blob/master/ scripts/generic/multi-bleu.perl} 
and ROUGE-l~\cite{lin-2004-rouge}.
%~\footnote{https://github.com/facebookresearch/ParlAI/blob/master/ parlai/core/metrics.py}%, and Unigram F1~\cite{dinan2018wizard}$.^{3}$

\noindent$\bullet$ \textit{Diversity}: the ratio of distinct uni/bi-grams in all generated texts, i.e., Distinct-1/2~\cite{li-etal-2016-diversity}. 

\noindent$\bullet$ \textit{Informativeness}: 
For WoW, we consider wikiF1~\cite{dinan2018wizard}, the overlapping F1 between the generated response and the grounded knowledge. For CRD, we calculate entity score (Ent.)~\cite{zhou2018commonsense}, the average number of entities per response. 
To further measure the likelihood of generating safe responses, we define \emph{\%safe}: the percentage of responses that contains ``I'm not sure'' or ``I don't know''.~\footnote{Upon manual inspection, we find that these two are the most common safe responses generated.} We also report the accuracy of knowledge selection (ACC) following \citet{zheng-etal-2020-difference}. 

%\noindent
%

We further perform human annotations by randomly sampling 
200/200/300/300 examples from WoW Test Seen/WoW Test Unseen/ CRD/DailyDialog, respectively. We recruit 5 annotators from a commercial annotation company to rate each response on a scale of 1-5 for its \emph{appropriateness}~\cite{zhang-etal-2020-grounded,zheng-etal-2020-difference} and  \emph{informativeness}~\cite{young2018augmenting,zhu-etal-2019-retrieval}. The former measures whether the topic of the response fits that of the utterance, while the latter evaluates whether a response provides new information. 
A response is scored 1 if it is not appropriate/informative at all, 3 if part of the response is appropriate/informative, 5 if it is highly related to utterance and context or it can provide rich information to deepen the discussion. 2 and 4 are for decision dilemmas.

%\paragraph{Configurations}
%Detailed model structures and training parameters are given in Appendix~\ref{app:config}. 

\section{Experiments}
We evaluate the performance of KI by comparing it with three sets of baselines:
\begin{enumerate}[wide=0.5\parindent,noitemsep, topsep=0.5pt]
    \item We first investigate the effectiveness and general applicability of KI by applying KI on conventional dialog models that are randomly initialized and trained with utterance-response pairs only.
    \item We then investigate whether KI can complement or even further improve the state-of-the-art KGD model's performance. %We then apply KI on state-of-the-art KGD models that are trained with utterance-response-knowledge triples.
    \item As discussed in \S\ref{sec:related}, although LMs differ from KI in many aspects, they also capture knowledge in their parameters. We thus compare KI with LMs to investigate its effectiveness in encouraging informative and appropriate responses. %Pre-trained language models that memorize a large amount of knowledge in their parameters. 
\end{enumerate}
%\bi{need to state here in each setting, what do u want to examine?}\zy{yes. plan to do this}
All model structures and training setups are given in Appendix~\ref{app:config}. 

\subsection{vs. Conventional Dialog Models}
\label{subsec:base_experiment}
%To investigate the effectiveness and general applicability of KI, 
We first deploy KI on two representative neural dialog models that do not directly condition on any external knowledge: 
(i) \textit{Seq2Seq}: a LSTM-based~\cite{hochreiter1997long} sequence-to-sequence model with the attention mechanism~\cite{bahdanau2014attention};
(ii) \textit{Transformer}~\cite{vaswani2017attention}: an encoder-decoder architecture relying solely on the attention mechanisms.

\paragraph{Effectiveness}

\begin{table*}[ht]
	\centering
	\resizebox{\textwidth}{!}{
		\begin{tabular}{c|l|cccccc|cccccc}
			\toprule
			\multicolumn{14}{l}{\textit{Setup 1: Models without externalized knowledge (trained with utterance-response pairs)}} \\
			\midrule
			\multirow{2}{*}{\textbf{Row}} &\multirow{2}{*}{\textbf{Model}} & \multicolumn{6}{c|}{\textbf{DailyDialog}} & \multicolumn{6}{c}{\textbf{CRD}} \\
 			& &\multicolumn{2}{c}{\textbf{PPL}} & \textbf{BLEU-4} & \textbf{ROUGE-l} &\textbf{Distinc-1/2} & \textbf{\%safe} &\textbf{PPL} & \textbf{Ent.} & \textbf{BLEU-4} & \textbf{ROUGE-l} &\textbf{Distinc-1/2} & \textbf{\%safe}\\
			\toprule
			1 & \textbf{Seq2Seq} & \multicolumn{2}{c}{28.94} & 3.84 & 14.22 & 2.85/11.74 & 2.50 &  55.54 & 1.32 & 2.59 & 10.58 & 1.13/4.47 & 41.81 \\ 
	        2 &\textbf{Seq2Seq+KI} & \multicolumn{2}{c}{29.35} & 4.65 & 14.64 & 3.36/14.10 & 2.70 & 47.32 & 2.26 & 2.90 & 11.13 & 1.86/7.37 & 35.08 \\ \hline
			3 &\textbf{Transformer} & \multicolumn{2}{c}{23.37} & 2.65 & 12.97 & 1.48/5.10 & 7.14 & 35.86 & 2.99 & 2.12 & 11.88 & 2.01/7.40 & 23.90 \\ 
			4 &\textbf{Transformer+KI} & \multicolumn{2}{c}{19.72} & 6.13 & 17.48 & 4.39/21.88 & 0.53 & 28.50 & 3.29 & 3.01 & 11.92 & 3.24/17.81 & 8.05\\ 
% 			5 &\textbf{Bert2Rnd} & \multicolumn{2}{c}{-} & - & - & -/- & - & - & - & - & - & -/- & - \\ 
% 			6 &\textbf{Ernie2Rnd} & \multicolumn{2}{c}{-} & - & - & -/- & - & - & - & - & - & -/- & - \\ 
			\midrule
			\multirow{2}{*}{\textbf{Row}} & \multirow{2}{*}{\textbf{Model}} & \multicolumn{6}{c|}{\textbf{WoW Test Seen}} & \multicolumn{6}{c}{\textbf{WoW Test Unseen}} \\ 
			& &\textbf{PPL} &\textbf{wikiF1} & \textbf{BLEU-4} & \textbf{ROUGE-l} &\textbf{Distinc-1/2} & \textbf{\%safe} &\textbf{PPL} & \textbf{wikiF1} & \textbf{BLEU-4} & \textbf{ROUGE-l} &\textbf{Distinc-1/2} & \textbf{\%safe}\\ 
			\toprule
			5 &\textbf{Seq2Seq} & 77.50 & 6.15 & 1.94 & 10.09 & 1.81/5.48 & 53.02 & 144.64 & 6.11 & 1.47 & 10.78 & 2.58/10.25 & 36.06 \\ 
			6 &\textbf{Seq2Seq+KI} & 67.69 & 9.59 & 2.25 & 12.45 & 4.99/17.32 & 36.24 & 122.46 & 7.09 & 1.62 & 11.23 & 3.12/12.05 & 37.98 \\ \hline
			7 &\textbf{Transformer} & 48.91 & 6.83 & 2.02 & 11.29 & 1.95/4.44 & 83.69 & 93.92 & 5.43 & 1.48 & 10.08 & 1.43/3.27 & 84.67 \\ 
			8 &\textbf{Transformer+KI} & 46.68 & 10.69 & 2.85 & 12.84 & 5.66/18.68 & 35.18 & 93.02 & 7.13 & 1.82 & 11.23 & 3.82/12.98 & 41.62  \\
% 			5 &\textbf{Bert2Rnd} & - & - & - & -/- & - & - & - & - & - & -/- & - \\ 
% 			6 &\textbf{Ernie2Rnd} & - & - & - & -/- & - & - & - & - & - & -/- & - \\ 
			\midrule
			\multicolumn{12}{l}{\textit{Setup 2: Models with externalized knowledge (trained with utterance-response-knowledge triples)}} \\
			\midrule
			\multirow{2}{*}{\textbf{Row}} & \multirow{2}{*}{\textbf{Model}} & \multicolumn{6}{c|}{\textbf{WoW Test Seen}} & \multicolumn{6}{c}{\textbf{WoW Test Unseen}} \\ 
			& &\textbf{ACC} &\textbf{wikiF1} & \textbf{BLEU-4} & \textbf{ROUGE-l} &\textbf{Distinc-1/2} & \textbf{\%safe} &\textbf{ACC} & \textbf{wikiF1} & \textbf{BLEU-4} & \textbf{ROUGE-l} &\textbf{Distinc-1/2} & \textbf{\%safe}\\ 
			\toprule
			9 &\textbf{DiffKS} & 25.30 & 67.06 & 5.73 & 17.48 & 9.61/37.29 & 5.10  & 19.72 & 64.77 & 4.60 & 15.75 & 3.83/12.15 &  7.36 \\
			10 &\textbf{DiffKS+KI} & 26.24 & 74.23 & 6.14 & 17.82 & 9.96/39.61 & 6.34 & 21.08 & 72.03 & 5.11 & 16.97 & 4.10/13.38 & 8.26 \\
			\bottomrule
		\end{tabular}
	}
	\caption{Automatic evaluation results for models with internalized knowledge (trained with utterance-response pairs), and models with externalized knowledge (trained with utterance-response-knowledge triples).  %We conduct significance tests between models with and without KI for BLEU (with bootstrap resampling~\cite{koehn-2004-statistical}), with $p$-value $< 0.01/0.05$ (denoted by */**).
	%\bi{why only sig test on bleu?}
	}
	\label{tab:main}
\end{table*}

As shown in Table~\ref{tab:main}'s Setup 1 (rows 1-8), dialog models with KI consistently outperform their counterparts without KI on almost all the metrics across the datasets used.
We want to point out the advantage of KI from two perspectives: 

\textit{(1) Promoting informativeness.}
We first observe that applying KI can significantly improve the wikiF1 and Ent. scores. 
Unlike KGD models that can generate informative responses by explicitly copying words from given knowledge, models discussed here are not provided with any external knowledge during testing (thus copy mechanism is not applicable for them).
This suggests that the improvement in informativeness should be attributed to the effectiveness of KI in injecting knowledge into models' parameters. The \textit{Info.} scores from human evaluation in Table~\ref{tab:main:human} can also substantiate our findings. 

\textit{(2) Promoting diversity and reducing occurrence of safe response.} 
Compared with the plain models,
models with KI can significantly improve the Distinc-1/2 scores on all the test sets (sometimes doubled, even tripled).
We also see a significant reduction of safe responses by the gap in \%safe scores. 
Those improvements are powered by the rich lexical knowledge used in KI (see Appendix~\ref{app:retrieval}).

\paragraph{Efficiency}

Besides the improvements in responses' quality, KI is also very efficient during inference. We report the decoding speed of Transformer and Transformer+KI in Table~\ref{tab:time}.
As we can see, KI does not incur any extra computation during inference. 

\subsection{vs. KGD}
\label{subsec:kgd}
%the state-of-the-art KGD model: to investigate whether introducing more fine-grained lexical knowledge can further improve the state-of-the-art KGD model's performance
We then apply KI on \textit{DiffKS}~\cite{zheng-etal-2020-difference}~\footnote{\url{github. com/chujiezheng/DiffKS}}: a state-of-the-art model  that uses a knowledge-aware decoder to generate a response based on utterance and the knowledge retrieved from a set of candidates. In the empirical study, DiffKS has outperformed many KGD models like CCM~\cite{zhou2018commonsense}~\footnote{Comparison with CCM is in Appendix~\ref{app:ccm}} and ITDD~\cite{li-etal-2019-incremental}. We enhance DiffKS by applying KI on its context encoder. The rest of the model remains unchanged. %We also compare to CCM in Appendix~\ref{app:ccm}. % In the empirical study, DiffKS has outperformed many KGD models like CCM~\cite{zhou2018commonsense} (see comparison in Appendix~\ref{app:ccm}) and ITDD~\cite{li-etal-2019-incremental}.

Table~\ref{tab:main} Rows 9-10 show that DiffKS with KI improves ACC over the plain DiffKS model.
The reason is that with the injection of token-level knowledge, DiffKS can better understand the utterance, which leads to more accurate knowledge selection and thus less noisy external knowledge.
As a result, we observe clear gains on overlapping-based metrics (BLEU and ROUGE). These results emphasize the importance of more fine-grained knowledge in KGD. Human evaluation results (Table~\ref{tab:main:human}) also suggest that KI can help KGD models in generating more informative and appropriate responses.

\begin{table*}[]
    \centering
    \resizebox{\textwidth}{!}{
    \begin{tabular}{l|cc|cc|cc|cc|l|cc|cc}
    \toprule
    \multirow{2}{*}{\textbf{Model}} & \multicolumn{2}{c|}{\textbf{DailyDialog}} & \multicolumn{2}{c|}{\textbf{CRD}} & \multicolumn{2}{c|}{\textbf{WoW Seen}} & \multicolumn{2}{c|}{\textbf{WoW Unseen}} & \multirow{2}{*}{\textbf{Model}} & \multicolumn{2}{c|}{\textbf{WoW Seen}} & \multicolumn{2}{c}{\textbf{WoW Unseen}}\\
    & \textbf{Appr.} & \textbf{Info.} & \textbf{Appr.} & \textbf{Info.} &\textbf{Appr.} & \textbf{Info.} & \textbf{Appr.} & \textbf{Info.} & & \textbf{Appr.} & \textbf{Info.} & \textbf{Appr.} & \textbf{Info.}\\ \hline
     \textbf{Transformer}    & 3.65(1.27) & 2.15(1.08) & 3.65(1.27) & 3.15(1.08) & 3.0(1.3) & 3.2(1.2) & 2.9(1.2) & 3.3(1.1) & \textbf{DiffKS}    & 3.6(1.0) & 3.6(1.0)& 3.7(0.9) & 3.7(1.0)\\
     \textbf{Transformer+KI} & 4.22(1.27) & 3.51(1.11) & 4.22(1.27) & 3.51(1.11) & 3.7(0.9) & 3.5(0.9) & 3.7(1.0) & 3.6(0.8) & \textbf{DiffKS+KI} & 3.9(0.9) & 4.0(0.9) & 4.0(0.9) & 4.2(0.8) \\
     \textbf{Human Response}          & 4.73(1.23) & 3.21(1.24) & 4.73(1.23) & 4.21(1.24) & 4.4(0.7) & 4.3(0.8) & 4.5(0.7) & 4.5(0.7) & \textbf{Human Response} & 4.5(0.7) & 4.3(0.8) & 4.4(0.8) & 4.2(0.8) \\
     \bottomrule
    \end{tabular}
    }
    \caption{Average of human annotations results on Appropriateness (Appr.) and Informativeness (Info.). Standard deviations are shown in the brackets.
    }
    \label{tab:main:human}
\end{table*}

\begin{table}[]
    \centering
    \resizebox{\linewidth}{!}{
    \begin{tabular}{l|ccc|ccc}
    \toprule
    \multirow{2}{*}{\textbf{Dataset}} &\multicolumn{3}{c}{\textbf{Transformer}} & \multicolumn{3}{c}{\textbf{Transformer+KI}} \\ 
     & \textbf{sent/s} & \textbf{tok/s} & \textbf{Time(s)} & \textbf{sent/s} & \textbf{tok/s} & \textbf{Time(s)}\\
    \midrule
    DailyDialog     & 215   & 2136 & 31.4 & 192 & 1980 & 35.1 \\
    CRD             & 158   & 5263 & 126.7& 184 & 4506 & 108.8\\
    WoW Seen        & 131   & 3397 & 25.9 & 133 & 2925 & 25.4 \\
    WoW Unseen      & 152   & 3943 & 22.4 & 140 & 3331 & 24.3\\
    \bottomrule
    \end{tabular}
    }
    \caption{Number of sentences/tokens decoded per second in testing and the total decoding time (in seconds).\label{tab:time}}
\end{table}

\subsection{vs. Pre-trained Language Models}
\label{sec:plms}

We follow previous practice~\cite{rothe2020leveraging} to replace the Transformer's encoder with LMs and keep the decoder the same as the \textit{Transformer} above.~\footnote{We keep the hidden state dimension of decoder consistent with the LMs to enable encoder-decoder attention.} We consider two baselines:
(i) \textit{Bert2Rnd}: Initializing Transformer's encoder with a pre-trained BERT, which has been shown capturing rich factual knowledge during pre-training~\cite{petroni2019language,jiang-etal-2020-know}.
(ii) \textit{Ernie2Rnd}: Initializing the encoder with ERNIE 2.0~\cite{sun2020ernie}, a knowledge-enhanced BERT which is pre-trained with novel objectives that injecting lexical, syntactic, and semantic knowledge into its parameters~\cite{zhang-etal-2019-ernie}.

From Table~\ref{tab:bert}, we see that parameters of LM-based models (Bert2Rnd and Ernie2Rnd) are more than three times than that of the Transformer baseline. But they do not seem to help improve informativeness (based on wikiF1, BLEU-4, and Info.) of responses. This indicates that although pre-trained LMs can encode knowledge in their parameters, eliciting the encoded knowledge for response generation is difficult when we only have utterance-response pairs for training. Another reason might be that previously learned knowledge is forgotten due to catastrophic forgetting~\cite{mccloskey1989catastrophic}. Comparing with knowledge-enhanced LMs, KI is more lightweight and more effective.% in improving the informativeness of responses (for encoder-decoder-based architectures). 

In addition, we observe that introducing LMs can significantly improve responses' diversity as KI does. However, according to Appr. metric and upon manual examination, we find that although the generated responses are diverse, they are often inconsistent with the context or hallucinating non-existing facts (e.g., "Yea, Canada is the largest country in the US."). These are known issues for LMs as discussed in~\citet{dou2021scarecrow,shuster2021retrieval,chen-etal-2020-kgpt}. 

We also apply KI on Bert2Rnd/Ernie2Rnd, but we do not observe significant improvements as when applied on randomly initialized models. This could be due to the fact that we implement KI using knowledge from Wikipedia, which is already part of LMs' training corpora. We leave it as future work to investigate how to use KI to elicit knowledge from LMs better (e.g., use adapters~\cite{xu2021retrieval} or prompt~\cite{liu2021pre}). 

\begin{table}[]
	\centering
	\resizebox{\linewidth}{!}{
		\begin{tabular}{l|c|cccccc}
			\toprule
			\textbf{Setting} &\textbf{\# Para} & \textbf{wikiF1} & \textbf{BLEU-4} &\textbf{Distinc-1/2} & \textbf{Info.} & \textbf{Appr.}  \\ 
			\toprule
			Transformer & 42.9M & 6.83 & 2.02 & 1.95/4.44 & 3.0(1.3) & 3.2(1.2) \\ \hline
			Bert2Rnd & 147.9M & 4.89 & 0.94 & 3.96/15.35 & 2.2(1.2) & 2.4(1.2) \\ % bert2share_lr1e-4 
			Ernie2Rnd & 147.9M & 5.15 & 0.91 & 3.95/19.73 &  2.2(1.1) & 2.3(1.2)  \\ % ernie2share 
			Transformer+KI & 43.2M & \textbf{11.25} & \textbf{2.85} & 5.66/18.68 & \textbf{3.7(0.9)} & \textbf{3.5(0.9)} \\ \hline
			Bert2Rnd+KI & 148.5M & 5.19 & 1.31 & \textbf{8.23/40.98} & 2.6(1.2) & 2.6(1.2) \\ % ki_bert2share_lr1e-4_m0.3
			Ernie2Rnd+KI & 148.5M & 5.02 & 1.19 & 5.01/21.27 & 2.3(1.1) & 2.4(1.2) \\ % outputs/wow/ki_ernie2share_lr1e-4_m0.3_vdp0.1/
			\bottomrule
		\end{tabular}
	}
	\caption{Results on LMs-based dialog generation.}
	\label{tab:bert}
\end{table}

%% file: 6-analysis.tex
\section{Method Analysis}

In this section, we perform an in-depth analysis to understand the effectiveness of KI.

\subsection{Working Principle of KI}
\label{sec:mechanism}
\begin{figure*}[ht]
    \centering
     \begin{minipage}{0.4\textwidth}
    \subfigure[Transformer\label{fig:emb_a}]{
    \includegraphics[width=\linewidth]{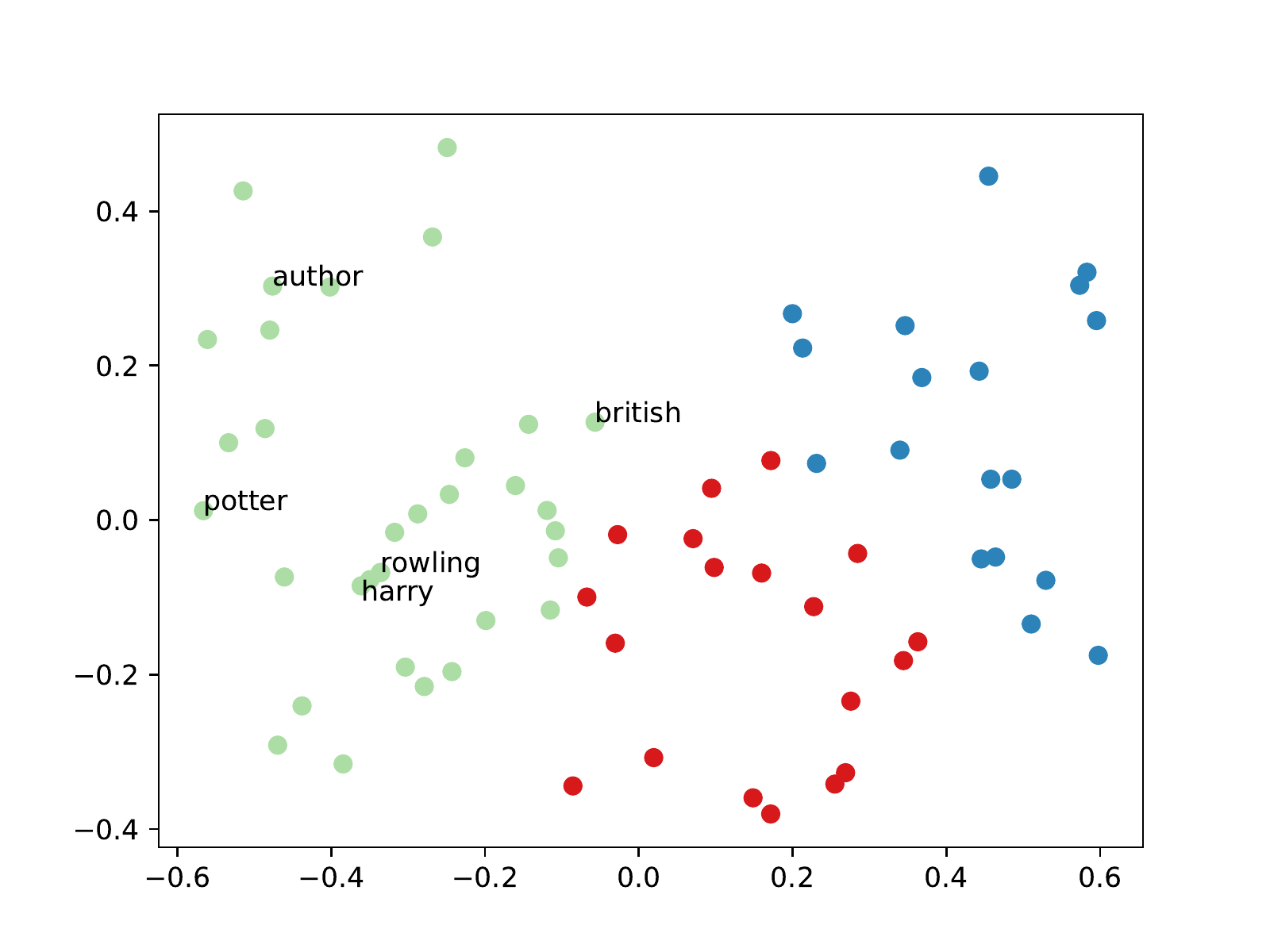}
    }
    \end{minipage}%
    \begin{minipage}{0.4\textwidth}
    \subfigure[Transformer+KI \label{fig:emb_b}]{
    \includegraphics[width=\linewidth]{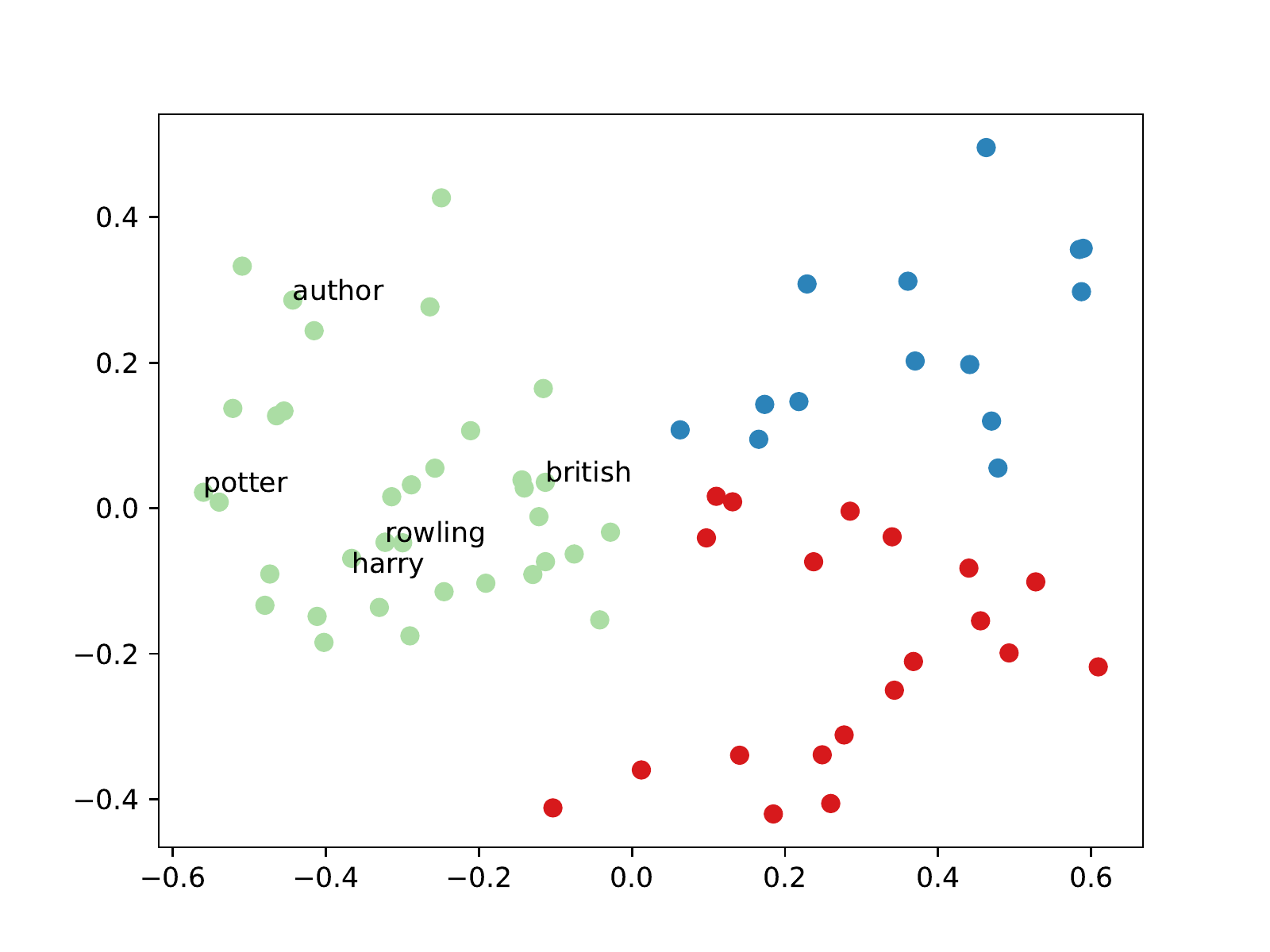}
    }
    \end{minipage}%
    \caption{Visualization of word embeddings learned by Transformer and Transformer+KI. We use words from two sources: 1)  lexical knowledge retrieved for \textit{Rowling}: ``J.K. Rowling is a British author and philanthropist.''  2) tokens from WoW that co-occur with ``Rowling'' in a sentence. Note that there is no co-occurrence of Rowling and British/author in WoW. All words are lower cased in the visualization. We use the K-means algorithm to group tokens into 3 clusters (shown in different colors).}
    \label{fig:embeddings}
\end{figure*}

We investigate the working principle of KI by visualizing the token embeddings learned on WoW. 
We use principal component analysis (PCA) to
map embeddings 
into a two-dimensional space 
as shown in Fig~\ref{fig:embeddings}. 
%We first note that this space is well separated with tokens of close semantic nearing each other (e.g., Harry and Rowling). 
Since there is no co-occurrence of British and Rowling in WoW, 
their embeddings learned by Transformer are distant (see Fig~\ref{fig:emb_a}). 
However, their embeddings learned by Transformer+KI (see Fig~\ref{fig:emb_b}) are much closer.
This is because KI injects lexical knowledge (i.e., a British author) into the embedding of Rowling. 
Specifically, 
the Euclidean distances between British and Rowling are 0.37 for Transformer and 0.22 for Transformer+KI, respectively. 
This observation sheds light on the working principle of KI: 
the contrastive learning objective shortens the embedding distance between a token and tokens from its lexical knowledge. Thus when decoding, if a token is predicted (e.g. Rowling), its relevant knowledge tokens (e.g., British) are likely to receive high probabilities and be selected in the following steps (see the J.K Rowling example in Fig~\ref{fig:intro_b}.
% an utterance token(e.g. Rowling) and tokens from relevant lexical knowledge (e.g., British). Thus when decoding, if a token is predicted, its related knowledge tokens are likely to receive high probabilities and be selected in the next step.

\subsection{Effectiveness of Token-level Knowledge}
\label{sec:token_analysis}

\begin{figure*}[ht]
    \centering
    \begin{minipage}{0.33\linewidth}
    \subfigure[BLEU-4\label{fig:ana_a}]{
    \includegraphics[width=\linewidth]{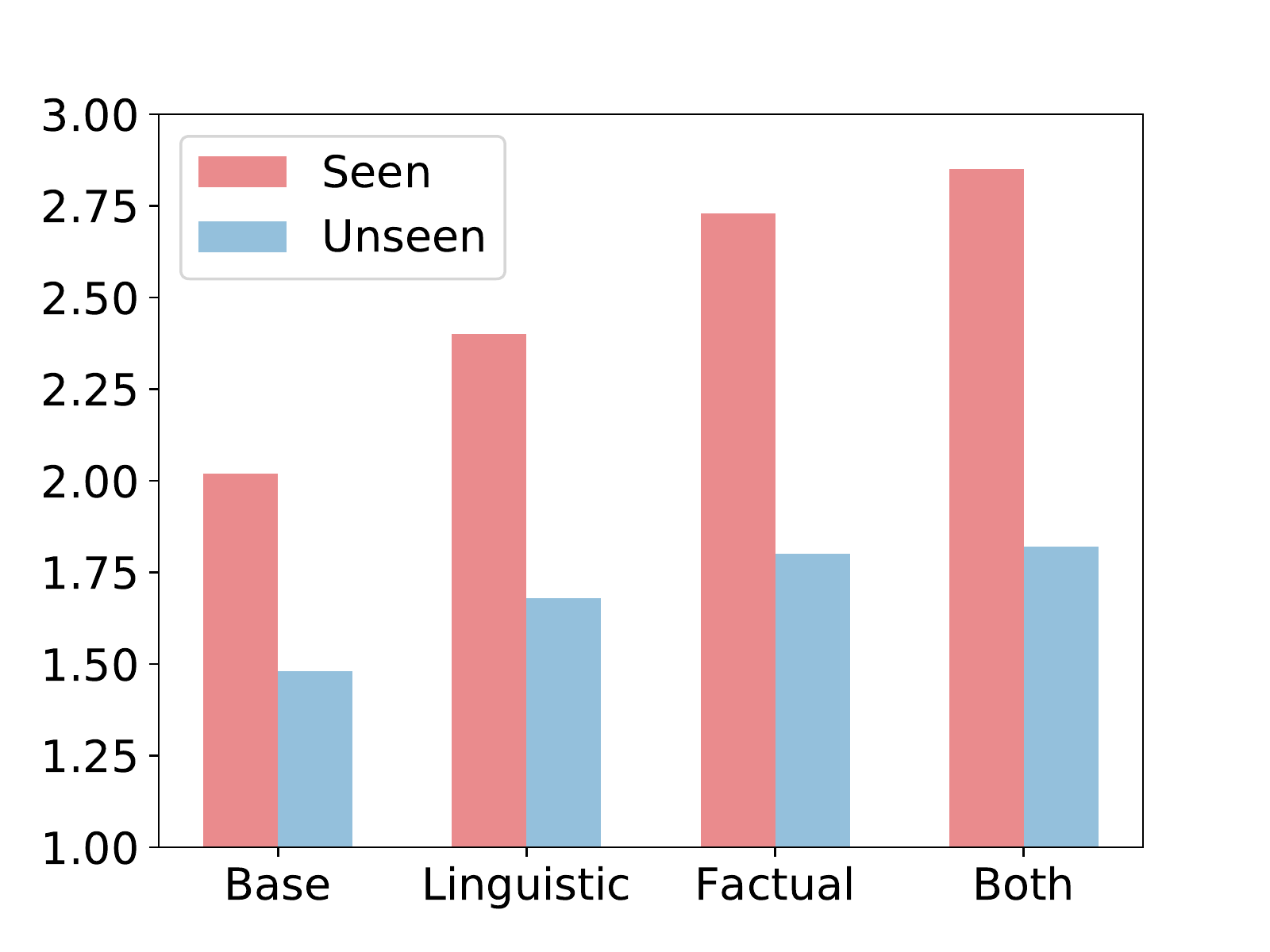}
    }
    \end{minipage}%
    \begin{minipage}{0.33\linewidth}
    \subfigure[ROUGE-l\label{fig:ana_b}]{
    \includegraphics[width=\linewidth]{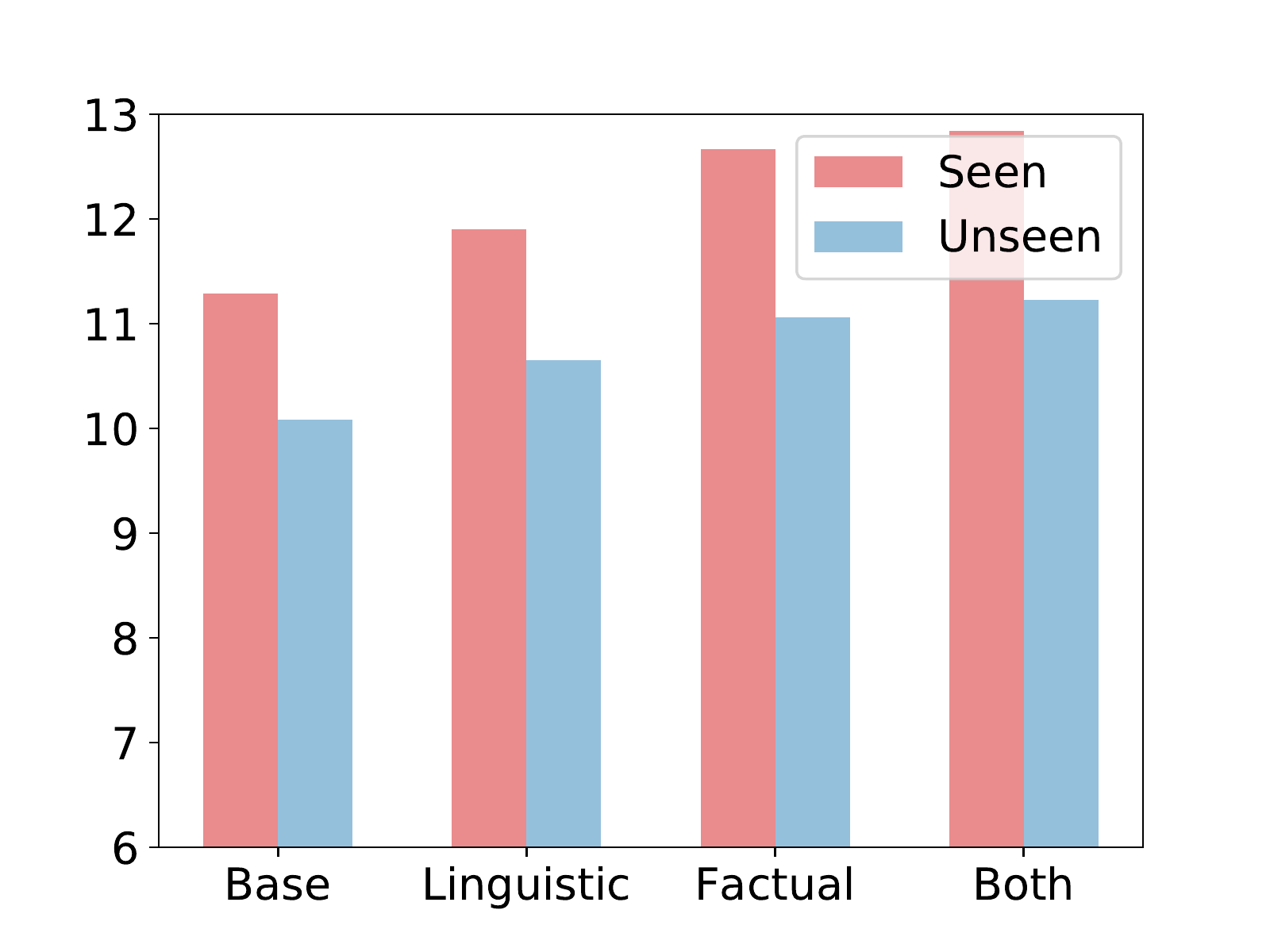}
    }
    \end{minipage}%
    \begin{minipage}{0.33\linewidth}
    \subfigure[Knowledge Coverage\label{fig:ana_c}]{
    \includegraphics[width=\linewidth]{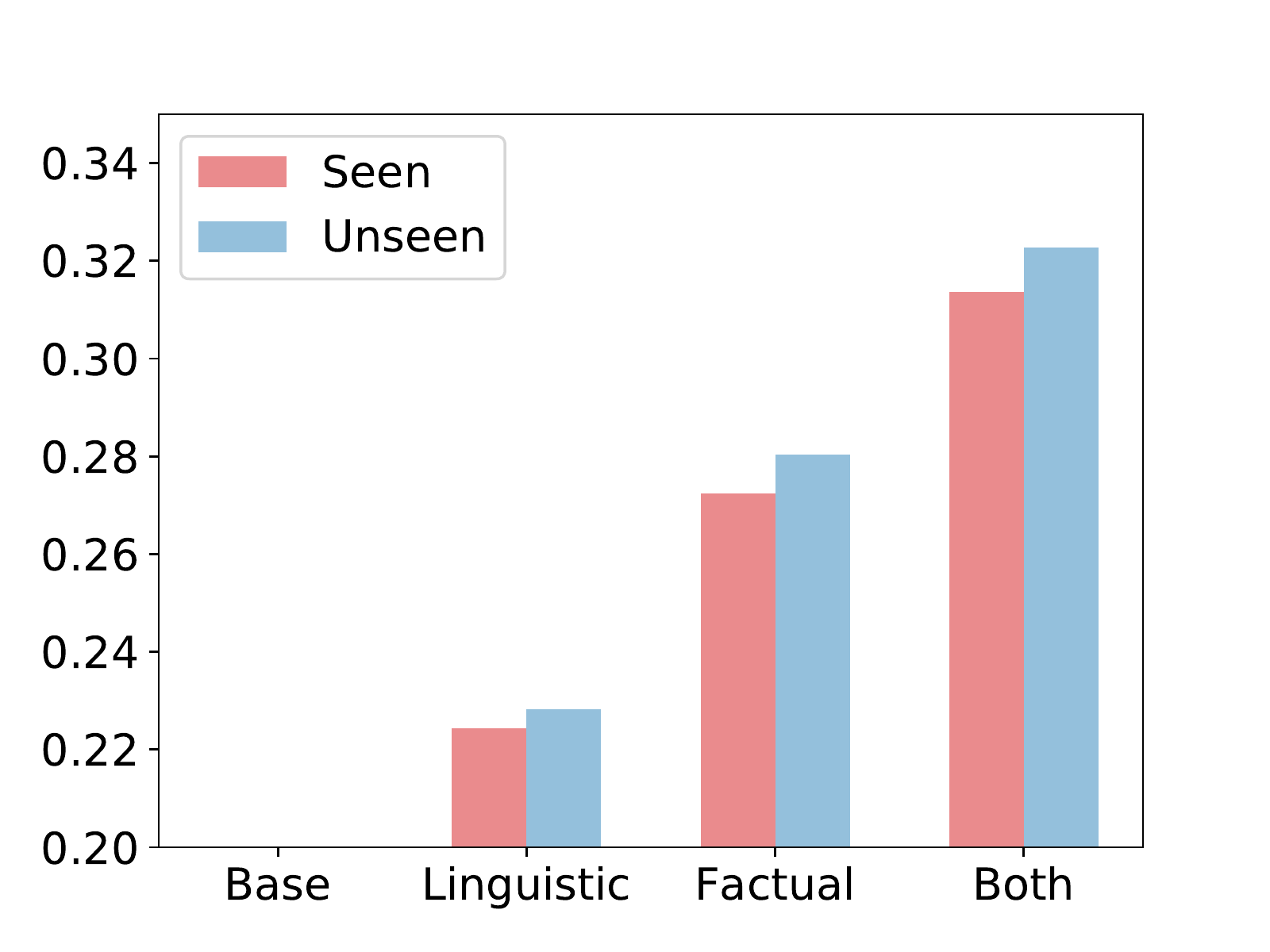}
    }
    \end{minipage}
    \caption{Automatic evaluation results on WoW Test Seen and Unseen. \textit{Base} is the Transformer baseline without KI. \textit{Both} is the Transformer+KI, with both linguistic and factual knowledge. \textit{Linguistic}/\textit{Factual} only considers linguistic/factual knowledge in KI, respectively.}
    \label{fig:token_analysis}
\end{figure*}

Firstly, we experiment with a model variant (denoted as Random),
which randomly assign knowledge to each utterance token. 
Results in Table~\ref{tab:variant} (Row 2) validate the effectiveness of the proposed token-knowledge retriever.

To further show the advantage of token-level knowledge, 
we consider a model variant in which we degenerate token-level KI to sentence-level by assigning all utterance tokens to a same lexical knowledge (we denote it as \emph{Sentence-level knowledge}). Given the lexical knowledge retrieved for each token in an utterance, the sentence-level knowledge is chosen as the most-frequent one among all token-level knowledge. %\li{totally lost in this sentence}  
The results are summarized in Table~\ref{tab:variant} (Row 3). 
%We find that use sentence-level knowledge still bringing improvement over Transformer baseline(Table~\ref{tab:main} row 7), this further prove the effectiveness of KI. \li{Why?totally lost.}
Note that token-level knowledge results in better performance than sentence-level knowledge.
This shows that fine-grained information is useful in promoting more informative and diverse responses.
%The results are expected as token-level KI exposes the model to more unique words, and based on our observation in Sec~\ref{sec:mechanism}, this lead to more diverse and informative responses.
%Although exposing model to more fine-grained knowledge is helpful, it does not mean that any randomly assigned knowledge to each token will be useful.

Lastly, we dive deep into the lexical knowledge retrieved to investigate which type of knowledge is most helpful in response generation. We classify a retrieved knowledge into two types: factual knowledge, which describes a real-world subject (e.g., knowledge about J.K Rowling), and is often associated with noun words in the utterance; linguistic knowledge, which explains the meaning of certain words (e.g., knowledge about donate, see Fig~\ref{fig:intro_b}, and is often associated with words except nouns.
We use part-of-speech (POS) tags to classify tokens and their associated knowledge.
We consider two model variants that only use factual/linguistic knowledge in KI respectively, denoted as \textit{factual} and \textit{linguistic}. In Fig~\ref{fig:token_analysis}, we compare these two model variants to a vanilla model without KI (denoted as \textit{base}),  and a full model that uses both knowledge (denoted as \textit{both}).
We find that injecting factual knowledge brings significant improvements on BLEU-4 and ROUGE-l. We also observe similar, albeit smaller improvements when equipping with linguistic knowledge. More interestingly, these two types of knowledge can complement one another to further improve the model performance. This emphasizes the need to consider non-factual knowledge in KGD, which is usually ignored in previous study. 
To understand what causes the difference between using factual and linguistic knowledge, we compute \emph{Knowledge Coverage}: the percentage of ground truth response tokens that have been recalled 
in the retrieved knowledge. 
As we can see from Fig~\ref{fig:ana_c}, factual knowledge is more helpful because people tend to respond based on knowledge related to subjects (usually nouns) appearing in the dialog. 

\begin{table}[]
	\centering
	\resizebox{\linewidth}{!}{
		\begin{tabular}{c|l|ccccc}
			\toprule
			\textbf{Row} &\textbf{Setting} & \textbf{wikiF1} & \textbf{BLEU-4} & \textbf{ROUGE-l} &\textbf{Distinc-1/2} & \textbf{\%safe} \\ 
			\toprule
			1 & \textbf{Token-level} & 11.25 & 2.85 & 12.84 & 5.66/18.68 & 35.18 \\
			2 & \textbf{Random} & 5.38 & 1.27 & 9.39 & 1.01/2.31 & 92.10 \\
			3 & \textbf{Sentence-level} & 8.41 & 2.31 & 11.72 & 2.98/7.77 & 66.32 \\
% 			3 & \textit{w/ non-nouns knowledge}  & 6.81/2.06 & 2.35 & 2.42/6.53 \\
% 			4 & \textit{w/ exact match} & 7.31/2.54 & 2.86 & 6.36/20.5\\
			\bottomrule
		\end{tabular}
	}
	\caption{Comparison of model variants for Transformer+KI, using different type of knowledge. Models are evaluated on WoW Test Seen.}
	\label{tab:variant}
\end{table}

\subsection{Case Study}
We show an example case in Appendix~\ref{app:case} to demonstrate how KI improves dialog generation and what the limitation is.

%% file: 9-appendix.tex
\clearpage
\appendix

\section{Dataset Statistics}
\label{app:dataset}
\begin{table}[]
    \centering
    \resizebox{\linewidth}{!}{
    \begin{tabular}{l|rrr}
    \toprule
     Dataset & Train & Valid & Test  \\ \hline
     WoW    &  166,787  & 17,715 & 8,715/8,782\\
     CRD    & 3,384,185 & 20,000 & 10, 000 \\
     DailyDialog & 54,889 & 6,005 & 5,700 \\ 
     \bottomrule
    \end{tabular}
    }
    \caption{Dataset statistics. WoW includes two test sets: Test Seen/Unseen, where the latter contains topics that never appear in Train and Valid set.}
    \label{tab:dataset}
\end{table}

\section{Implementation Details}
\label{app:config}
%During training of the token-level knowledge retrieval model~\footnote{https://github.com/airsplay/vokenization}, we fine-tune two \textit{bert-base-uncased} model separately. 

The vocabulary size for DailyDialog/WoW/CRD is 14,696/22,168/22,512, respectively, with sentences tokenized using BERT's tokenizer provided by Transformers~\cite{wolf-etal-2020-transformers}. For Seq2Seq and Transformer, we use a shared vocabulary between the encoder and the decoder. 
In Seq2Seq, we adopt a 2-layer bidirectional LSTM as the encoder and an unidirectional one as the decoder. The hidden size is set to 256, with a dropout probability of 0.3. 
The Transformer we used has 6 encoder/decoder layers. The dimensions of the input layer, output layer, and inner feed-forward layer are set to 512, 512, and 1,024, respectively. The number of attention heads is set to 4.

We use Adam with $\beta_1=0.9$, $\beta_2=0.98$ for model optimization and start training with a warm-up phase where we linearly increase the learning rate from $10^{-7}$ to 0.005. After that we decay the learning rate proportional to the number of updates. Each training batch contains at most 4,096 source/target tokens. We early-stop the training if validation loss does not improve over ten epochs. We perform beam search with a beam size of 5.
The $\lambda$ (see Eq~\ref{eq:ki}) is set to 1 in all our evaluation. 

For Bert2Rnd and Ernie2Rnd, we initialize the Transformer's encoder with the pre-trained LMs using the Transformers~\cite{wolf-etal-2020-transformers} and keep the decoder the same as above. Note that due to the exist of encoder-decoder attention, we modify the dimensions of input/output layer to 768 to be compatible with BERT (\textit{bert-base-uncased}) and ERNIE (\textit{nghuyong\/ernie-2.0-en}). We share the embeddings between encoder and decoder. Models are learned with Adam optimizer with $\beta_1=0.9$, $\beta_2=0.98$. Learning rate is set to 1e4 with a linear scheduler. Each training batch contains 128 samples. The LMs are fine-tuned together with the decoder. We also experimented with LMs frozen, but this generally works worse.

\section{Analysis of Token-level Knowledge Retrieval}
\label{app:retrieval}
Since our retrieval component is based on the contextualized representations (see \S~\ref{sec:retrieval-encoder}), the same token can be aligned to different knowledge when it occurs in different contexts. As the supporting evidence, in Table~\ref{tab:retrieval_statistics}, we report the averaged number of knowledge items associated with each token. In Table~\ref{tab:knowledge_items}, we show an example of the same token being aligned to different knowledge items when giving different contexts. In addition, our approach exposes each dialog sentence to very diverse knowledge items. The rich lexical knowledge, both at the token-level and sentence-level, is the key to KI's good performance.  

\begin{table}[]
    \centering
    \resizebox{\linewidth}{!}{
    \begin{tabular}{l|ccc}
    \toprule
     Number of knowledge items  & WoW & DailyDialog & CRD \\ \hline
     per token    & 30 & 26 & 38 \\
     per sentence & 15 & 12 & 9 \\
    \bottomrule
    \end{tabular}
    }
    \caption{Averaged number of knowledge items associated with each token/sentence.}
    \label{tab:retrieval_statistics}
\end{table}

\begin{table}[]
\small
    \centering
    %\resizebox{\linewidth}{!}{
    \begin{tabular}{p{0.9\linewidth}}
    \toprule
    \textbf{Context:} one of our favorite books is the wonderful \textcolor{mypink}{wizard} of oz by author l . frank ba \#\#um and published in 1900 ! \\ \hline
    \textbf{Knowledge:}   The Wonderful Wizard of Oz is an American children's novel written by author L. Frank Baum and illustrated by W. W. Denslow, originally published by the George M. Hill Company in May 1900. \\ \hline
    \textbf{Context:}   it ' s about a young \textcolor{mypink}{wizard} at hog \#\#wart \#\#s , right ? \\ \hline
    \textbf{Knowledge:}  The book follows Harry Potter, a young wizard, in his third year at Hogwarts School of Witchcraft and Wizardry. \\  
    \bottomrule
    \end{tabular}
  %  }
    \caption{An example case from WoW. Given different contexts, the token \textit{wizard} is aligned to different knowledge items.}
    \label{tab:knowledge_items}
\end{table}

We further conduct an ablation study to investigate the effectiveness of two additional retrieval strategies: stopword masking and exact matching (\S~\ref{sec:biases}).
We remove each strategy and keep the other unchanged.
The results are presented in Table~\ref{tab:ablation}. As we can see,
both strategies are useful for generating appropriate (based on PPL, BLEU-4, and ROUGE-l), informative (based on WikiF1), and diversified (based on Distinc-1/2) responses.
%\bi{the last sentence does not directly describe the results.}
\begin{table}[]
	\centering
	\resizebox{\linewidth}{!}{
		\begin{tabular}{c|l|ccccc}
			\toprule
			\textbf{Row} &\textbf{Setting} & \textbf{wikiF1} & \textbf{BLEU-4} & \textbf{ROUGE-l} &\textbf{Distinc-1/2} & \textbf{PPL}  \\ 
			\toprule
			1 & \textbf{Transformer+KI} & 11.25 & 2.85 & 12.84 & 5.66/18.68 & 46.68 \\
			2 & \textit{w\/o stopwords masking} & 5.23 & 2.63 & 12.68 & 5.48/21.74 & 57.27 \\
			3 & \textit{w\/o exact matching} & 10.74 & 2.54 & 11.99 & 4.75/15.27 & 47.65 \\
% 			3 & \textit{w/ non-nouns knowledge}  & 6.81/2.06 & 2.35 & 2.42/6.53 \\
% 			4 & \textit{w/ exact match} & 7.31/2.54 & 2.86 & 6.36/20.5\\
			\bottomrule
		\end{tabular}
	}
	\caption{Retrieval strategy ablation results on WoW Test Seen.}
	\label{tab:ablation}
\end{table}

\section{Comparison with CCM}
\label{app:ccm}
Similar to KI, CCM augments dialog corpora with token-level commonsense knowledge. In each encoding and decoding step, CCM explicitly uses the retrieved commonsense knowledge triples by concatenating their representations with the token representation. As existing KGD models, CCM also requires extra knowledge as input during both training and inference. Training CCM on the CRD dataset takes about a week on one Titan X GPU. The comparison of model performance is shown in Table~\ref{tab:ccm}. As we can see, there is a significant gap between CCM and Transformer+KI. Thus in \S\ref{subsec:kgd}, we consider applying KI on a more state-of-the-art and recent KGD model: DiffKS. 

\begin{table}[]
    \centering
    \begin{tabular}{l|cc}
    \toprule
    \textbf{Model}     &  \textbf{PPL}  & \textbf{Ent.} \\ \hline
    CCM  &  39.18 & 1.18 \\
    Transformer+KI & 28.50 & 3.29 \\
    \bottomrule
    \end{tabular}
    
    \caption{Automatic evaluation on CRD. Numbers of CCM are taken from their paper.}
    \label{tab:ccm}
\end{table}

\section{Case Study}
\label{app:case}
\begin{table}[]
\small
    \centering
    %\resizebox{\linewidth}{!}{
    \begin{tabular}{p{0.9\linewidth}}
    \toprule
    % $d_1$  & \textbf{Utterance:} i like chocolate what of you \\
    %         & \textbf{Reference:} i love chocolate ! it is a sweet , brown food prepared from cacao seeds . \\
    %         & \textbf{Transformer:} i love chocolate ! \\ 
    %         & \textbf{Transformer+KI:} i love chocolate . it is made in the form of a liquid , paste , or in a block \\
    %         & \textbf{Knowledge for ``chocolate'':} Chocolate is a preparation of roasted and ground cacao seeds that is made in the form of a liquid, paste, or in a block, which may also be used as a flavoring ingredient in other foods. \\
    % \midrule 
    \textbf{Context:} \\
    SpeakerA: I like Dylan's Bars, do you? \\
    SpeakerB: Yes Dylan's Candy Bar is my favorite boutique candy store. \\ \hline
    \textbf{Utterance:}   They have everything! I just love it. \\ \hline
    \textbf{Gold Response:}   Yes Ralph Lauren's daughter Dylan Lauren owns them. \\ \hline
    \textbf{Transformer:}  I'm not sure , but I do know that they have been around for a long time! \\  \hline
    \textbf{Transformer+KI:}  I love their chocolate chip cookies! They're actually the second \textcolor{mypink}{largest} \textcolor{mypink}{candy} company \textcolor{mypink}{in the world}! \\ 
    \textbf{Knowledge for Dylan's:}  Lauren was inspired to create the store, which is asserted to be the "\textcolor{mypink}{largest} unique \textcolor{mypink}{candy} store \textcolor{mypink}{in the world}", by the Roald Dahl story of Willy Wonka the Chocolate Factory \\
    % daughter 158517 A daughter is a female offspring- a girl or woman in relation to her parents .
    % candy 24939 Candy , also called sweets ( British English ) or lollies ( Australian English , New Zealand English ) , is a confection that features sugar as a principal ingredient .
    \textbf{Knowledge for like:} In English, the word like has a very flexible range of uses, ranging from conventional to non-standard.\\
    \bottomrule
    \end{tabular}
  %  }Overlapped tokens between generated responses and the retrieved knowledge are in pink.
    \caption{An example case from WoW Seen.}
    \label{tab:cases}
\end{table}

We show an example case in Table~\ref{tab:cases} to demonstrate how KI improves dialog generation and what the limitation is. 
%For each dialog, we show the utterance, the context of the conversation (if there is any), response generated by Transformer and Transformer+KI, and token-level knowledge for particular token(s) in utterance.
%As we can see, for the, a Transformer baseline simply copy parts of the utterance as response and does not provide new information or try to engage the partner in the conversation. By internalizing knowledge about ``chocolate'' into Transformer, we find that Transformer+KI produces a much informative response. Despite different from the reference response provided, we found the response to be coherent and informative. Note that our response has string overlapping with the retrieved knowledge sentences, despite the fact that the model does not explicitly ``see'' the knowledge sentence during inference. This support our claim that KI can inject knowledge into models' parameter. 
From the generated results, Transformer returns a vacuous response, as it has no idea on what ``Dylan's Candy Bar'' is. 
However, Transformer+KI, which perceives the knowledge about ``Dylan's Candy Bar'' during training, gives a much more informative response. 
Meanwhile, we further observe some inaccuracy during the knowledge transfer (``largest'' becomes ``second largest''). We take this as an interesting future work.

\section{Comparison with BART}
In \S~\ref{sec:plms}, we observe that KI can outperform models whose encoders are initialized with pre-trained BERT or ERNIE. Here we dive deeper to compare KI with a fully pre-trained seq2seq model: BART~\cite{lewis-etal-2020-bart}. BART has demonstrated superior performance on conditional language generation, including translation, summarization, and dialogue response generation. We start from the BART-base checkpoint~\footnote{https://huggingface.co/facebook/bart-base}. We fine-tune the model for five epochs with a learning rate of 3e-5. We do not report PPL since these two models use different tokenization methods. As we can see from Table~\ref{tab:bart}, by introducing only a few extra parameters and computation, KI can significantly boost the Transformer's performance. Although a pre-trained BART model can generate slightly more diverse responses than Transformer+KI (higher \textit{Distinc-1/2}), these generated responses are often inconsistent with the input (lower \textit{BLEU-4/ROUGE-l}) or less informative (lower \textit{WikiF1}).

\begin{table*}[ht]
	\centering
	\resizebox{\textwidth}{!}{
		\begin{tabular}{c|l|cccc|cccc}
			\toprule
			\multirow{2}{*}{\textbf{Row}} &\multirow{2}{*}{\textbf{Model}} & \multicolumn{4}{c|}{\textbf{DailyDialog}} & \multicolumn{4}{c}{\textbf{CRD}} \\
 			& & \textbf{BLEU-4} & \textbf{ROUGE-l} &\textbf{Distinc-1/2} & \textbf{\%safe} & \textbf{BLEU-4} & \textbf{ROUGE-l} &\textbf{Distinc-1/2} & \textbf{\%safe}\\
			\toprule
			1 &\textbf{Transformer} & 2.65 & 12.97 & 1.48/5.10 & 7.14 & 2.12 & 11.88 & 2.01/7.40 & 23.90 \\ 
			2 &\textbf{Transformer+KI} & 6.13 & 17.48 & 4.39/21.88 & 0.53 & 3.01 & 11.92 & 3.24/17.81 & 8.05\\ 
			3 &\textbf{BART-base} & 0.65 & 13.40 & 4.95/19.47 & 5.51  & 0.48 & 11.60 & 5.03/26.89 & 7.40 \\ 
			\midrule
			\multirow{2}{*}{\textbf{Row}} & \multirow{2}{*}{\textbf{Model}} & \multicolumn{4}{c|}{\textbf{WoW Test Seen}} & \multicolumn{4}{c}{\textbf{WoW Test Unseen}} \\ 
			& & \textbf{WikiF1} & \textbf{BLEU-4/ROUGE-l} &\textbf{Distinc-1/2} & \textbf{\%safe} & \textbf{WikiF1} & \textbf{BLEU-4/ROUGE-l} &\textbf{Distinc-1/2} & \textbf{\%safe}\\ 
			\toprule
			4 &\textbf{Transformer} & 6.83 & 2.02/11.29 & 1.95/4.44 & 83.69 & 5.43 & 1.48/10.08 & 1.43/3.27 & 84.67 \\ 
			5 &\textbf{Transformer+KI} & 10.69 & 2.85/12.84 & 5.66/18.68 & 35.18 & 7.13 & 1.82/11.23 & 3.82/12.98 & 41.62  \\
			6 &\textbf{BART-base} & 8.85 & 1.99/11.30 & 6.43/24.91 & 15.79 & 6.51 & 1.65/11.85 & 5.16/20.90 & 15.82 \\ 
			\bottomrule
		\end{tabular}
	}
	\caption{Automatic evaluation results for Transformer+KI and BART-base.
	}
	\label{tab:bart}
\end{table*}